\documentclass[conference]{IEEEtran}
\IEEEoverridecommandlockouts
\usepackage{cite}
\usepackage{amsmath,amssymb,amsfonts}
\usepackage{algorithmic}
\usepackage{graphicx}
\usepackage{textcomp}
\usepackage{xcolor}
\usepackage{lineno,hyperref}
\usepackage{mathabx}
\usepackage{subcaption}
\usepackage{todonotes}
\usepackage{multirow}
\usepackage{float}

\DeclareMathOperator*{\avg}{avg}

\newcommand{\fperf}{\ensuremath{\mathit{perf}}}

\usepackage{comment}

\def\BibTeX{{\rm B\kern-.05em{\sc i\kern-.025em b}\kern-.08em
    T\kern-.1667em\lower.7ex\hbox{E}\kern-.125emX}}
    
\begin{document}

% \title{Continuous performance evaluation for business process outcome monitoring}
\title{Measuring the Stability of Process Outcome Predictions in Online Settings}
% \title{Stability! Something very important to know about your online prediction framework...}
% \title{Is my model stable? A new perspective towards evaluating online process outcome prediction frameworks...}
% The instability of online process outcome prediction frameworks 
% STOMPPOT: STability Of Model for Predictive Process Outcome over Time.

\author{\IEEEauthorblockN{Suhwan Lee}
\IEEEauthorblockA{
Utrecht University\\
Utrecht, The Netherlands\\
Email: s.lee@uu.nl}
\and
\IEEEauthorblockN{Marco Comuzzi}
\IEEEauthorblockA{
Ulsan National Institute of Science and Technology\\
Ulsan, Republic of Korea\\
Email: mcomuzzi@unist.ac.kr}\\ 
\IEEEauthorblockN{Hajo A. Reijers}
\IEEEauthorblockA{
Utrecht University\\
Utrecht, The Netherlands\\
Email: h.a.reijers@uu.nl}
\and 
\IEEEauthorblockN{Xixi Lu}
\IEEEauthorblockA{
Utrecht University\\
Utrecht, The Netherlands\\
Email: x.lu@uu.nl}
}

\maketitle
\thispagestyle{plain} %force page number to appear
\pagestyle{plain} %force page number to appear

\begin{abstract}

Predictive Process Monitoring aims to forecast the future progress of process instances using historical event data. 
As predictive process monitoring is increasingly applied in online settings to enable timely interventions, evaluating the performance of the underlying models becomes crucial for ensuring their consistency and reliability over time. This is especially important in high risk business scenarios where incorrect predictions may have severe consequences.
%
%Diagnostic information regarding model performance in an online setting not only supports better decision-making, but also facilitates effective assessment of potential risks. 
%
%For example, in high-risk business scenarios, unstable predictive models with significant decreases in performance should not be used, despite a relatively higher average performance over time. 
%
However, predictive models are currently usually evaluated using a single, aggregated value or a time-series visualization, which makes it challenging to assess their performance and, specifically, their stability over time.
This paper proposes an evaluation framework for assessing the stability of models for online predictive process monitoring. The framework introduces four performance meta-measures: the frequency of significant performance drops, the magnitude of such drops, the recovery rate, and the volatility of performance. To validate this framework, we applied it to two artificial and two real-world event logs. 
The results demonstrate that these meta-measures 
facilitate the comparison and selection of predictive models for different risk-taking scenarios.
Such insights are of particular value to enhance decision-making in dynamic business environments.
\end{abstract}

\begin{IEEEkeywords}
predictive monitoring, process outcome, event stream, performance evaluation
\end{IEEEkeywords}

\section{Introduction}

% Highlights the increasing application of PPM in online settings for timely intervention

The pervasiveness of information systems in business process execution, empowered in particular by IoT systems~\cite{schonig2018integrated}, allows event data to be logged and become available for process analytics in real-time. This has prompted the process mining discipline to look into the \emph{online} realization of typical use cases, such as predictive process monitoring~\cite{pauwels2021incremental}. 
% In an online setting, event data is logged continuously and made instantly available for analysis. 
In \emph{online} settings, an event log is a stream of events becoming available for analysis as soon as they are logged. 

The use of predictive models in real-life settings requires accurate performance evaluation to facilitate effective decision-making and mitigate potential risks while adapting to changing business environments. 
For example, in business scenarios where the predictions may be used to make decisions with \emph{high risk} (e.g., selecting the proper treatment of a patient in a hospital),
incorrect predictions may lead to suboptimal decision-making, potentially with severe consequences.
% In such high-risk business scenarios, prediction models that have unstable performance or significant decreases in performance are less reliable and should, therefore, be opposed, despite the fact that the prediction model may have a relatively higher average performance over time.
%
% In high-risk business scenarios, 
% it is important to keep consistency in the performance to minimize the potential risk of an incorrect prediction. 
% For instance, in high-risk business scenarios like a patient diagnosis in hospitals, 
In high-risk scenarios, the predictive model should perform stably over time with minimal deviation on the performance metrics from recent evaluations, considering that each decision carries significant risk. %In such high-risk business scenarios, 
Predictive models that have an unstable performance or lose performance over time are to be avoided, despite the fact that they may have a relatively higher average performance over time. In online settings in particular, business users may also want to prioritize models that demonstrate quick recovery to their accustomed performance levels. For example, if the process can be dynamically changed, the predictive model should be capable of efficiently incorporating newly arrived data, enabling rapid learning and adaptation.

Existing performance evaluation approaches often use a single aggregated value or time-series visualization. While they provide valuable performance diagnostics, they do not provide measures for the long-term performance of the model or its capability to adapt to dynamic changes. These are both concerns that relate to the \textit{stability} of the process monitoring framework. Let us assume two example plots of performance: Scenario 1 and Scenario 3 in Fig.~\ref{fig:matrix}. Despite having the same average accuracy, each model can be analyzed from various perspectives regarding their performance fluctuation. For instance, the accuracy of Scenario 1 shows higher volatility compared to Scenario 3, indicating \emph{unstable} performance. Scenario 1 can quickly adapt and learn from newly observed data, allowing it to recover its accustomed performance levels. This example shows that it is not trivial to determine which model delivers the most desired level of stability.

% Introduce a novel evaluation framework designed to assess the stability of performance in online predictive models for process outcome monitoring
% Describes four performance meta-measures included in the framework

This paper proposes an evaluation framework with four meta-measures of performance for online process outcome predictive monitoring, which can help to evaluate the performance stability of an online predictive model in a balanced way. 
% In  using four meta-measures of performance, which evaluate the performance stability of an online predictive classifier that may be helpful in real-world scenarios. 
These are: (1) the frequency at which the performance of a model tends to drop from its average, (2) the average magnitude of such performance drops, (3) the speed at which the average performance levels can be recovered, and (4) the volatility of the performance level over time. Given any of the traditional confusion matrix-based measures for predictive model performance, the proposed meta-measures can generate concrete measures to assess the performance stability of an online outcome-oriented predictive monitoring model.  

% Mention a validation of the evaluation framework using artificial and real-life event logs
% Highlights the results obtained from applying the framework in facilitating the comparison and selection of suitable classifiers for different risk-staking scenarios
% Emphasizes that by considering performance stability, organizations can ensure reliable predictions and make informed decisions in dynamic business environments.

The proposed performance evaluation framework is evaluated on two artificial event logs and two publicly available real-life event logs. The results show that the performance measures obtained can be used to assess the stability of the performance of these predictive models and support the selection of models depending on the scenario in which it is applied. 

 % \textbf{[TODO@Suhwan, please check the entire paper: classifier versus prediction model versus predictive model]}
 % In addition,  also to infer important insights regarding the behavior recorded in the stream of events.

The paper is organized as follows. The research problem and motivation are discussed in the next section. Section~\ref{sec:related} introduces  related work. Section~\ref{sec:prelimin} provides preliminaries, while the performance metrics are presented in Section~\ref{sec:performance_evaluation}. The empirical analysis is reported in Section~\ref{sec:empirical_analysis}, while conclusions are drawn in Section~\ref{sec:conclusion}.

\section{Problem Description and Motivation}
\label{sec:problem_motivation}

% We motivate the need for the proposed performance meta-measures in Section~\ref{sec:motivation}. Then, after having provided some required concepts in Section~\ref{sec:concepts}, the proposed performance meta-measures are defined in Section~\ref{sec:defs}.

% \subsection{Motivation}
% \label{sec:motivation}

%An accepted classification in process mining is the one distinguishing between the so-called \emph{lasagna} and \emph{spaghetti} business processes. The former tend to involve repeatable and possibly standardised operations with few predictable exceptions. Hence, they are modelled by relatively simple and regular business process models. The latter tend to involve chaotic and often unpredictable operations, with a higher chance of generating unanticipated exceptions. The procedural process models describing such processes tend to be complex and often unusable in practice. 

To highlight the need for more nuanced performance metrics in outcome-oriented predictive process monitoring, we introduce a classification of business scenarios along two dimensions: the \emph{frequency} of the decisions taken using a predictive monitoring model and the \emph{risk} associated with these decisions. The frequency dimension refers to the number of decisions that must be taken in a given time unit during the execution of a process with the support of a predictive model. The risk dimension refers to the cost associated with taking a wrong decision during the execution of a process case with the support of a predictive model.

\begin{figure}[h]
    \centering
    \includegraphics[width=\linewidth]{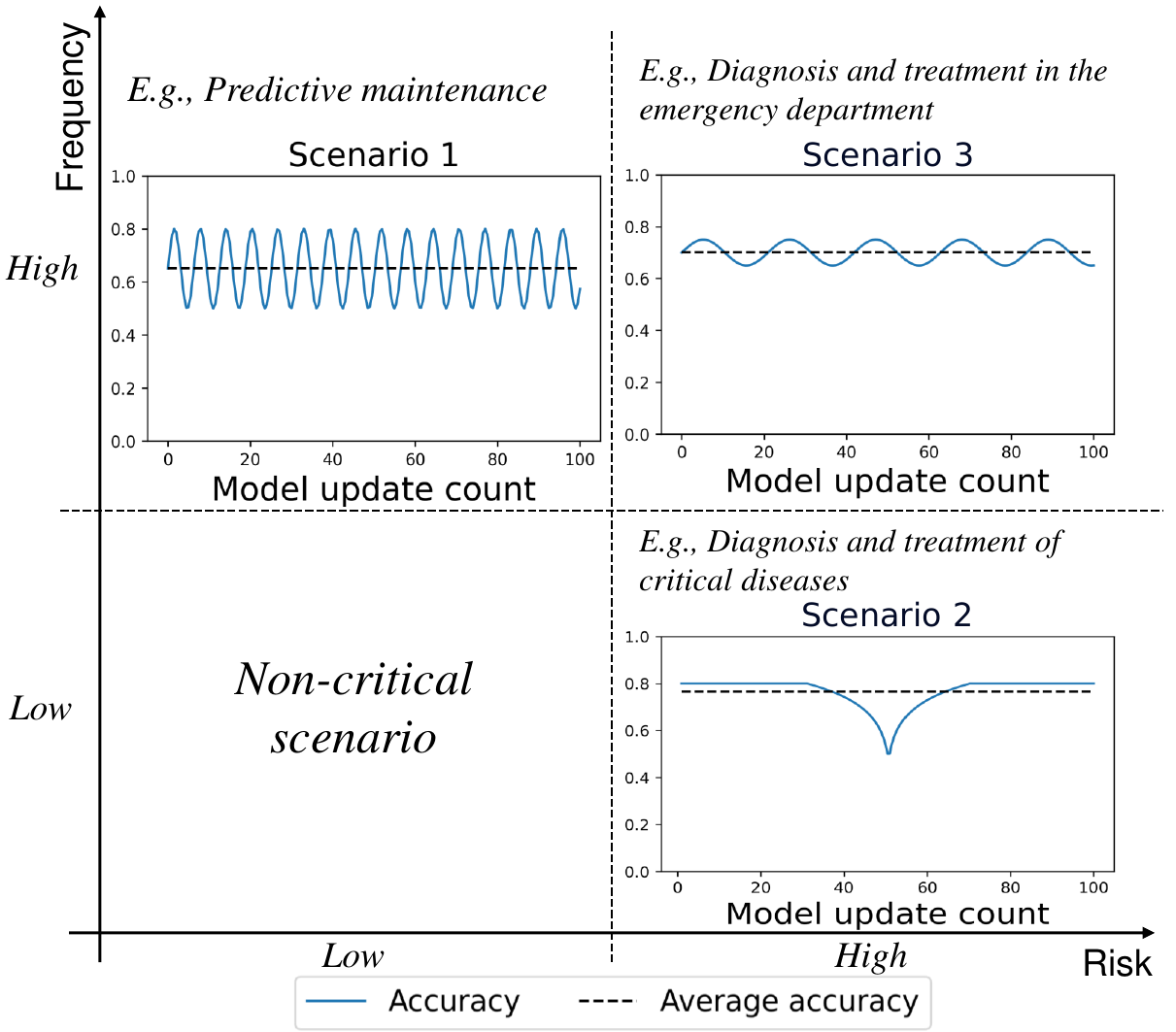}
    \caption{Scenarios for motivating the performance meta-measures.}
    \label{fig:matrix}
\end{figure}

Considering these two dimensions, business scenarios can be classified in a 4-cell matrix resulting from the combination of the values \emph{low} and \emph{high} for each dimension.

\begin{itemize}

\item \emph{Scenario 1 - High Frequency, Low Risk}: These are critical scenarios in which, while not being particularly risky, decisions have to be taken frequently. Because decisions have to be taken frequently in these scenarios, it is important to use predictive models with a stable outcome, i.e., for which the performance is on average acceptable and relatively steady in time. 
A typical example of this scenario is predictive maintenance of a large fleet of mechanical installations (e.g., elevators or other machinery). Online models in predictive maintenance are relevant because the machinery operating conditions may change abruptly over time and, therefore, it is advisable to use predictive models that can be continuously updated by new monitoring data. Still, predictive maintenance decisions are taken routinely and can be very frequent, particularly in scenarios where the machinery can suddenly degrade and it is therefore fundamental to have a means to anticipate such a degradation effectively and timely. 

\item \emph{Scenario 2 - Low Frequency, High Risk}:  These are critical scenarios in which, while not being particularly regular or frequent, incorrect decisions may result in excessive costs. In these types of scenarios, it is important to have online models with a good performance, i.e., for which the performance drop is never too sharp, and that, even when the performance drops, can recover to an acceptable level. 
A typical example of this scenario is the diagnosis and treatment of critical diseases like pancreatic cancer. Online models in diagnosis and treatment are relevant because the conditions under which such decisions are taken may vary dramatically over time, e.g., patient load at a hospital, availability of specialized expertise and treatment options, vital signs of patients, etc. Moreover, incorrect decisions are very risky, because they may result in bad outcomes for patients, loss of reputation, and financial loss. Therefore, it is important to equip decision-makers in such scenarios with predictive models that are highly reliable and stable. 
% Such prediction model may take time to retrain
% able to quickly recover to high performance values in reaction to a performance drop.

\item \emph{Scenario 3 - High Frequency, High Risk}:  There are highly critical scenarios that combine the challenges of the two cases discussed in Scenario 1 and Scenario 2. A typical example of this scenario is the diagnosis and treatment in the emergency department. 

\item \emph{Non-critical scenario}: These are non-critical scenarios in which decisions do not have to be taken often during a process case and neither are these decisions too risky, i.e., wrong decisions are generally recoverable and do not entail high costs.
\end{itemize}

%Decision makers may act under resource constraints, so they may have to choose whether to have a model that is, for instance, more accurate or quicker at recovery

%Decision makers often train different models, none of which are optimal, so they need to choose a model, that's were the performance metrics introduce in this paper are useful.

A straightforward way to adjust the performance measurement in the case of streaming data is to apply a traditional measure, e.g., accuracy or F1, to a sliding window of samples. This certainly gives insights into the model performance variation over time. However, it does not directly measure other performance aspects relevant to the scenarios of Fig.~\ref{fig:matrix}, like the speed at which the performance may recover to an acceptable level, or the frequency at which the performance of the model falls below a certain threshold. To address these concerns, it is crucial to devise additional performance measures to capture those aspects. In this paper, we propose four meta-measures. Given a traditional performance measure obtained from the analysis of a confusion matrix (e.g., accuracy or F-measures), meta-measures allow us to derive a new performance measure suitable for the evaluation of the stability of online outcome-oriented predictive process monitoring models. The first two meta-measures address the needs of high-frequency online business scenarios, whereas the other two are introduced to address the requirements of high-risk online business scenarios:

\begin{enumerate}

\item \emph{Frequency of relevant performance drops}, which captures the rate at which the performance of an online predictive model drops below a certain level considered acceptable. 

\item \emph{Volatility of the performance}, which captures the volatility over time of the performance of an online predictive model. 

\item \emph{Magnitude of performance drop}, which captures how much the performance of a model normally drops from its expected acceptable level. 

\item \emph{Recovery rate}, which measures the speed at which an online model normally recovers from a performance drop. 

\end{enumerate}

The performance measures derived from the application of the meta-measures can be used to compare competing online outcome-oriented predictive monitoring models. Decision makers often face resource constraints, requiring them to choose between a model that is, for example, more consistently accurate (with minimal relevant performance drops) or one that can quickly recover to an acceptable performance level after a drop occurs. More in general, decision makers are likely to train different predictive models in a given scenario. The performance measures outlined in this paper will help decisions regarding which model is more suitable in a specific business scenario.

\section{Related work}
\label{sec:related}

In this section, we discuss the performance evaluation approaches in online predictive process monitoring techniques, listed in Table~\ref{tab:related_works_table}. We categorize existing approaches based on two dimensions: (1) the evaluation approach and (2) the evaluation level. We provide detailed discussions for each category. Then, our contributions in relation to related works are presented. 

% %%%%%%%%%%%%%%%%%%%%%%%%%%%%%%%%%%%%%
% Aggregated value evaluation %
% %%%%%%%%%%%%%%%%%%%%%%%%%%%%%%%%%%%%%

The classical approach for evaluating model performance involves using aggregated performance metrics. In the case of conducting a \emph{prefix length level} analysis, an online predictive model can be assessed based on median accuracy. For instance, Maisenbacher and Weidlich~\cite{maisenbacher2017handling} have introduced the use of incremental classifiers to handle event streams, with the goal of developing outcome predictive models capable of adapting to concept drift. They evaluate their approach by comparing the median accuracy of the model under different concept drift scenarios across various prefix lengths.

Regarding \emph{event log level} analysis, overall prefix lengths are used to calculate the average accuracy for model evaluation. Pauwels and Calders~\cite{pauwels2021incremental} have introduced a deep neural network with an incremental learning strategy that can be updated over time to predict the next activity in an event stream. They measure the performance of their methods under different model training scenarios by calculating the average accuracy on all updates. Baier et al.~\cite{baier2020handling} have investigated the challenge of determining the optimal data selection points for retraining an offline predictive model when concept drift occurs in an event stream. The data selection points are analyzed by comparing the average accuracy of the entire data set.  Other than accuracy, the average F-score is used for the model evaluation. Di Francescomarino et al. ~\cite{di2018incremental} evaluate incremental classifiers in the context of process outcome prediction. They evaluated the models at various prefix lengths, combining the outcomes into a single value, and compared the aggregated results to a baseline for performance comparison, utilizing both average F-score and accuracy as evaluation metrics.

% %%%%%%%%%%%%%%%%%%%%%%%%%%%%%%%%%%%%%
% Time-series visualization %
% %%%%%%%%%%%%%%%%%%%%%%%%%%%%%%%%%%%%%

In the case of the evaluation approach for \emph{time-series visualization}, the prediction results obtained are often represented in a graph where the x-axis corresponds to the frequency of model updates. This visualization allows for the assessment of performance changes over time. In some cases, prediction approaches are evaluated not only based on a single aggregated value but also by examining the visualized representation of the results~\cite{maisenbacher2017handling, pauwels2021incremental, baier2020handling}. However, the visualization is only used to compare their proposed methods to the baseline models. 

Specially for \emph{event log level} performance visualization, Lee et al.~\cite{lee2022continuous} have introduced an evaluation framework for online outcome process prediction. They specifically focus on real-time model performance by analyzing the collected prediction results across entire prefix lengths. The framework utilizes average accuracy values from each prefix to generate a graph, allowing for the observation of performance changes over time. However, it primarily emphasizes the overall performance of the model over time, without explicitly considering the discovered performance fluctuations in relation to stability.

In summary, the existing evaluation approaches are not specifically designed for streaming event logs, and they do not assess the model's fluctuation in performance as it is updated. Using a single aggregated value for assessment may not capture the performance fluctuations accurately. In the case of time-series visualization, it has mainly focused on highlighting the outperforming predictive model. In addition, the discussion of model performance consistency along with model updates has been limited. In this paper, we propose meta-metrics for online predictive process outcome monitoring to measure the stability of performance over time.

\begin{table*}[]
\centering
\caption{Comparison of performance evaluation approaches in online predictive process monitoring}
\label{tab:related_works_table}
\resizebox{0.8\linewidth}{!}{%
\begin{tabular}{c|l|c|c|c|c|c}
\hline
\textbf{\begin{tabular}[c]{@{}c@{}}Evaluation \\ approach\end{tabular}} &
  \multicolumn{1}{c|}{\textbf{\begin{tabular}[c]{@{}c@{}}Evaluation\\  level\end{tabular}}} &
  \begin{tabular}[c]{@{}c@{}}Maisenbacher \\ and Weidlich~\cite{maisenbacher2017handling}\end{tabular} &
  \begin{tabular}[c]{@{}c@{}}Pauwels \\ and Calders~\cite{pauwels2021incremental}\end{tabular} &
  Baier et al.~\cite{baier2020handling} &
  \begin{tabular}[c]{@{}c@{}}Di Francescomarino\\  et al.~\cite{di2018incremental}\end{tabular} &
  Lee et al.~\cite{lee2022continuous} \\ \hline
\multirow{2}{*}{\begin{tabular}[c]{@{}c@{}}Aggregated \\ value\end{tabular}}         & Prefix length & v &   &   &   &   \\ \cline{2-7} 
                                                                                     & Event log     &   & v & v & v &   \\ \hline
\multirow{2}{*}{\begin{tabular}[c]{@{}c@{}}Time-series\\ visualization\end{tabular}} & Prefix length & v &   &   &   &   \\ \cline{2-7} 
                                                                                     & Event log     &   & v & v &   & v \\ \hline
\end{tabular}%
}
\end{table*}

\section{Preliminaries}
\label{sec:prelimin}

\subsection{Notations}
Given the first $n$ positive natural numbers $\mathbb{N}^+_n$ and a target set $S$, a sequence $s$ is a function $s:\mathbb{N}^+_n \to S$ mapping integer indexes to the elements of $S$.  Given a set of activity labels $A$, the domain of timestamps, and a set of $I$ attribute domains $D_i$, we define the set of event attributes as $E=A \times \mathbb{N}^+ \times [D_1 \times \ldots \times D_i \times \ldots \times D_I ] $.  A trace $\sigma$ is a sequence of $n$ events $\sigma:\mathbb{N}^+_n \to E$. We denote with $\mathcal{E}$ the event universe, with $\mathcal{E}= E \times J$, where $J$ is a set of possible case ids, and with $\mathcal{E}^*$ the universe of sequences of events. An event stream is an infinite sequence $\Psi: \mathbb{N}^+ \to \mathcal{E}$.
%@XIXI: fix 

For simplicity, we write events as $e_{k,j}$, where $k$ indicates their position in a trace and traces as $\sigma_j = \langle e_{1,j}, \ldots, e_{i,j}, \ldots, e_{N_j,j} \rangle$, where $N_j$ is the number of events in the trace $\sigma_j$. The function $t:\mathcal{E} \to \mathbb{N}^+$ returns the timestamp of an event. The prefix function $\mathit{pref}: \mathcal{E}^* \times \mathbb{N}^+ \nrightarrow \mathcal{E}^* $ is a partial function that returns the first $p$ events of a trace, i.e., $\mathit{pref}(\sigma_j,p)=\langle e_{1,j}, \ldots, e_{p,j} \rangle$, with $p \leq N_j$. Note that, for the evaluation, event streams are generated from event logs in which multiple events may have the same timestamp. For the events that have the same timestamp, we assume that the ordering of the events in an event log reflects their true ordering and use this order in the stream to calculate prefixes.

A trace $\sigma$ is associated with a binary outcome label and, without loss of generality, we assume that the value of this label becomes known with the last event $e_{N_j, j}$ of a trace. Therefore, we define a labeling function as a partial function $y: E \nrightarrow \{0,1\}$, which returns the label of a trace in correspondence with its last event. For clarity and with an abuse of notation, we denote the label of a trace $\sigma_j$ as $y_j$.

A sequence encoder is a function $f$, with $f: \mathcal{E}^* \to \mathcal{X}_1 \times \ldots \times \mathcal{X}_w \times \ldots \times \mathcal{X}_W$ mapping a prefix onto a set of features defined in the domains $\mathcal{X}_w$. A process outcome prediction model $m$ is a function $\hat{y}: \mathcal{X}_1 \times \ldots \times \mathcal{X}_w \times \ldots \times \mathcal{X}_W \to \{0,1\}$ mapping an encoded prefix onto its predicted label. 
%\todo{Maybe moving the above text to preliminaries?}
The prefixes may be divided into separate buckets. A different predictive model may be maintained (trained/tested) for each bucket of prefixes. In this work, we consider prefix-length bucketing~\cite{leontjeva2016complex} of traces, which is commonly adopted in the literature: a different predictive model $m_k$ is trained and tested using a set of prefixes of length $k=1,\ldots, K$, where the maximum prefix $K$ may vary for each event log. Thus, we define an outcome prediction framework $f$ as a collection of outcome predictive models $m_k$, that is, $f=\{ m_k \}_{k=1,\ldots, K}$. 
Note that the performance evaluation framework proposed in this paper does not depend on the method used to encode or bucket the prefixes.  

Next, we propose our methods to evaluate the performance of an online outcome prediction framework $f$.

% \subsection{Preliminaries for the meta-measures definitions}
\subsection{Moving Windows and Prediction Framework}
\label{sec:concepts}

While receiving a stream of events, a prediction framework $f$ can be updated, e.g., every time a new label (or a sufficient number thereof) is received. Most importantly, when a new label is received, a prediction framework can be re-evaluated. This is the typical \emph{continuous evaluation} scenario that we consider in this paper. 

Let $t_1$, $t_2$, $\cdots$, $t_k$ represent the points in time when the corresponding outcome prediction frameworks $f^{t_1}$, $f^{t_2}$, $\cdots$, $f^{t_k}$ are evaluated. We use a moving window approach, where the window size is denoted by $l$. This moving window indexes the last $l$ labels received, which corresponds to the most recent $l$ completed cases. To clarify, we use $W_{t_1}$, $W_{t_2}$, $\cdots$, $W_{t_k}$ to represent the sets of the last $l$ completed cases at time $t_1$, $t_2$, $\cdots$, $t_k$, respectively. %(Note that the cases in the windows may overlap). 

We introduce a performance measure $\theta$, which can be any metric like accuracy, error rate, F-score, AUC, or others. This measure takes a prediction framework $f^{t_i}$ and a set of completed cases $W_{t_i}$ to compute the performance of $f^{t_i}$, denoted as $\theta(f^{t_i}, W_{t_i}) = p_{t_i} \in [0,1]$. We extend the use of $\theta$ to calculate a sequence of performance values associated with the sequence of moving windows. Formally, let $f^* = \langle f^{t_1},f^{t_2}, \cdots, f^{t_k}\rangle$ be the sequence of prediction frameworks and $W^* = \langle W_{t_1}, W_{t_2}, \cdots, W_{t_k}\rangle$ the sequence of sets of completed cases at $t_1, t_2, 
\cdots t_k$, respectively; we overload $\theta( f^*, W^* ) = \langle\theta(f^{t_1}, W_{t_1}), \theta(f^{t_2}, W_{t_2}), \cdots, \theta(f^{t_k}, W_{t_k})  \rangle = \langle p_{t_1}, p_{t_2}, \cdots, p_{t_k}\rangle$. 
% where $\theta(i) = p_i = \theta(\hat{Y}(W_i))$. 
%
For example, assuming $\theta$ computes the AUC, and $\theta = \langle 0.75, 0.65, 0.85, ..., \rangle$, then $0.75$ is the AUC of the prediction framework $f^{t_1}$ on $W_{t_1}$, $0.65$ is the AUC of $f^{t_2}$ on $W_{t_2}$,  etc.

%\subsection{A sequence of performance values and significant drops}

\section{Continuous performance evaluation}
\label{sec:performance_evaluation}
Traditionally, to compare two online prediction frameworks, one tends to compute an aggregated measure of $\theta$ (e.g., the average of AUC). In this paper, we propose the four meta-measures of the performance measures based on the \emph{moving average} of $\theta$ to quantify the stability. 

In the following, we briefly recap the moving average, the moving standard deviation, and the lower-bound and upper-bound of the moving average over $\theta$, which we use to define the four performance meta-measures later. 
First, given a sequence of performance values $\theta(f^*, W^*) = \langle p_{t_1}, p_{t_2}, \cdots, p_{t_k}\rangle$, and $M \in \mathbb{N}^+$ a natural number, the \emph{moving average} $\mathit{MA}_M( \langle p_{t_1}, p_{t_2}, \cdots, p_{t_k}\rangle) = \langle ma_{t_1}, ma_{t_2}, \cdots, ma_{t_k}\rangle$ where $ma_{t_i} = \frac{1}{M} \cdot \sum_{i=k-M+1}^{k} p_{t_i}$ if $k>M$; otherwise, $ma_{t_i} = \frac{1}{k} \cdot \sum_{i=1}^{k} p_{t_i}$

We now consider the \emph{moving standard deviation} of the moving average and use that to define the significant changes in the performance sequence. 
% In particular, we refer to the negative changes, i.e., the ones where the performance falls below an acceptable level defined using the moving average and standard deviation of a sequence, as \emph{drops}. 
%
To avoid confusion, we use $\varphi_{t_i}$ to denote the standard deviation associated with $ma_{t_i}$ at $t_i$. Let $\theta(f^*, W^*)  = \langle p_{t_1}, \cdots, p_{t_k} \rangle = p^*$ be the sequence of performance measures, and $\mathit{MA}_M(p^*)= \langle \mathit{ma}_{t_1}, \cdots, \mathit{ma}_{t_k} \rangle$ the corresponding sequence of moving average, we compute the moving standard deviation $\varphi_M(p^*) = \langle \varphi_{t_1}, \cdots, \varphi_{t_k} \rangle$ at ${t_k}$ where $\varphi_{t_i} =  \sqrt{\frac{\sum_{j = i - M+1}^i (p_{t_j} - ma_{t_i})^2}{M}}$ if $i > M$; otherwise, $\varphi_{t_i} =  \sqrt{\frac{\sum_{j=1}^i (p_{t_j} - ma_{t_i})^2}{i}}$.

Let $\mathit{MA}_M(p^*) =  \langle ma_{t_1}, ma_{t_2}, \cdots, ma_{t_k}\rangle$ be the moving averages and $\varphi_M(p^*) = \langle \varphi_{t_1}, \cdots, \varphi_{t_k} \rangle$ the standard deviations, we define an upper-bound $up(ma_{t_i})$ at $t_i$ as $up(ma_{t_i}) = ma_{t_i} + \varphi_{t_i}$. Similarly, a lower-bound $lb(ma_{t_i}) = ma_{t_i} - \varphi_{t_i}$.
% We overload the functions $up$ and $lb$ to respectively return the sequence of upper-bounds of the moving band $up(\langle \mathit{ma}_1, \cdots, \mathit{ma}_m\rangle) = \langle up(\mathit{ma}_1), \cdots, up(\mathit{ma}_m)\rangle $) and the sequence of lower-bounds of the moving band $lb(\langle \mathit{ma}_1, \cdots, \mathit{ma}_m\rangle) = \langle lb(\mathit{ma}_1), \cdots, lb(\mathit{ma}_m)\rangle $. 
When a performance value $p_{t_i}$ is below the lower-bound $\mathit{lb(ma_{t_i})}$, we say this $p_{t_i}$ is a \emph{drop point} $d_i$, i.e., $d_i = p_{t_i}$. For example, assume that the performance values $p_2^* = \langle ..., 0.75, 0.65, 0.5, ...\rangle$, and the moving average $\mathit{MA}_M(p^*) = \langle ..., 0.7, 0.69, 0.66, ...\rangle$ and the moving standard deviation $\varphi_M(p^*) = \langle ..., 0.03, 0.03, 0.04, ...\rangle$. Then, the performance values $0.65$ and $0.5$ in $\theta_2$ are two drop points, since they are below the lower-bounds (0.66 and 0.62). 
% Comment XL: a figure needed?
% In the plot at the bottom of Fig.~\ref{fig:moving-numeric}, the blue band highlights the lower and upper bounds. Moreover, the significant points $p_{t_i}$ either above $up(ma_{t_i}$ and below $lb(ma_{t_i})$ are highlighted using red dots.
%(A drop point is highlighted as a RED dot in Fig.~\ref{todo}.)

%\subsubsection{Drop points and significant drops}
Using the drop points, we define a \emph{significant drop} $D = \langle p_{t_u}, \cdots, p_{t_v} \rangle$ as a sequence of \emph{consecutive} drop points. 
Formally, let $\theta(f^*, W^*) = \langle p_{t_1}, p_{t_2}, \cdots, p_{t_k}\rangle = p^*$ be the sequence of performance results, and $\mathit{MA}_M(p^*) = \langle ma_{t_1}, ma_{t_2}, \cdots, ma_{t_m} \rangle$ its sequence of moving average, we define a \emph{significant drop} $D = \langle p_{t_u}, \cdots, p_{t_v} \rangle$ as a sequence of consecutive drop points, such that
\begin{itemize}
    \item (1) for all $ t_u \leq t_i \leq t_v$, the performance $p_{t_i} \in \theta$ at $t_i$ is below the lower-bound $lb(ma_{t_i})$, i.e., $p_{t_i} < lb(ma_{t_i})$
    \item (2) $t_1 \leq t_u \leq t_v \leq t_m$, i.e., it is a subsequence of $\theta$
    \item (3) the previous point is not a drop point (i.e., either $p_{t_{u-1}} \geq lb(ma_{t_{u-1}})$ or $u = 1$ ), and
    \item (4) the next point is not a drop point (i.e.,  either $p_{v+1} \geq lb(ma_{v+1})$ or $v = m$).
\end{itemize}

Taking the previous example of $\theta_2$, the two consecutive significant drop points constitute a significant drop $D = \langle 0.65, 0.5 \rangle$.
We use $\mathbb{D}(\theta) = \{D_1, D_2, ... \}$ to refer to the set of all significant drops of a performance sequence.

\subsection{The four meta-measures}
\label{sec:defs}

% In this case, the lower-bound defined using the moving standard deviation acts as a buffer for the tolerated level of new performance points to the average and denoted it as $\sigma_{\fperf}$. The observed performance that is over the tolerated level is a drop $d$. 

% The new performance measurement $\fperf_{rt}(pof)_{i+1}$ that is deviated to the moving average is defined as follows:

% \[
% d = \{d \in D\vert \lvert \fperf_{rt}(pof)_{i+1} - MA_{\fperf} \rvert > \sigma_{\fperf}\}
% \]

Having defined the sequence of performance values, the moving average, and the moving standard deviation, including the lower bounds and upper bounds, we have defined the significant drops. Based on these characteristics of the performance values, we propose 4 meta-measures to continuously monitor the performance and provide performance diagnostics as discussed in Section~\ref{sec:problem_motivation}.
%calculate time volatility of the performance in various perspective.

\subsubsection{Frequency of significant performance drop}

% Number of drop points / Number of all points

The first meta-measure to analyze the volatility of the performance is the frequency of significant drops. This measure is designed to answer the question \emph{``How often the model performance significantly deviated from (below) the moving average?"} The measure is defined as the number of observed significant drops. Given the sequence of performance values $\theta = \langle p_1, ..., p_n \rangle$ and the drops $\mathcal{D}(\theta) = \{D_1, ..., D_m\}$, the (normalized) frequency $\mathbb{F}$ of significant performance drops is defined as follows:

% \[
% \mathbb{F} = \frac{\sum_{D_i \in \mathbb{D}(\theta)} \rvert D_i\lvert}{\rvert \theta \lvert} 
% \]

\[
\mathbb{F} = \lvert \mathbb{D}(\theta) \rvert 
\]

When the frequency of the significant drop is low, the prediction framework is stable and does not have a sudden decrease in its performance.

\subsubsection{Volatility of the performance}
The next meta-measure is the volatility of the performance, which answers the question \emph{``How volatile are the performance values of the prediction framework?"} The standard deviation from the moving average of the moving window is used to define this meta-measure. Given the sequence of the standard deviations $\langle \varphi_1, ..., \varphi_n \rangle$, the average of the sequence of standard deviations $\mathbb{V}_\fperf$ is defined as follows:

\[
\mathbb{V}_\fperf = \frac{\sum_{i=1}^{n}\varphi_{i}}{n}
\]

When the value of this measure is low, the variation of the performance values is low, thus the performance of the prediction framework is stable (or, in other words, less volatile).

% \textcolor{red}{volatility of the performance : $\overline{\varphi...}$}

\subsubsection{Magnitude of performance drop}
This meta-measure answers the question \emph{``How large is the absolute value of the performance drops of the framework?"} The magnitude of performance drop is the difference between a drop point $p_{i}$ and the moving average $ma_i$, i.e., $\lvert p_i - ma_i \rvert$. 
%
% \[
% \mathbb{M} = \{\lvert p_{i}-ma_{i} \rvert \vert D \in \mathbb{D}(\theta)\, p_{i} \in D\}
% \]
%
% From the collected $\mathbb{M}$ the severity of the deviated performance is assessed by the average and the maximum of performance drop. 
We then define the maximum magnitude $\mathbb{M}_{max}$ and the average magnitude $\mathbb{M}_{avg}$ of performance drop as follows:

\[
\mathbb{M}_{max} = \max_{D \in \mathbb{D}(\theta), p_i \in D}{ \lvert p_{i} - ma_{i} \rvert }
\]

\[
\mathbb{M}_{avg} = \avg_{D \in \mathbb{D}(\theta), p_i \in D}{ \lvert p_{i} - ma_{i} \rvert }
\]

When the maximum magnitude or the average magnitude of the drop points is low (e.g., close to 0.1), then the absolute decrease in the prediction performance is small, despite the drops being significant.

% \textcolor{red}{maximum of the performance drops: \[\max_{D \in \mathbb{D}(\theta), p_i \in D}{p_i - ma_i}\]}

% \textcolor{red}{average of the performance drops: \[\avg_{D \in \mathbb{D}(\theta)} \,\,\,\, \max_{p_i \in D}{p_i - ma_i}\]}

\subsubsection{Recovery rate}
The recovery rate meta-measure aims at measuring the speed of recovery of a prediction framework from significant drops to an acceptable performance level. This meta-measure is designed to answer the question \emph{``How quickly can the model recover from a sudden change in the prediction ability?"} The recovery rate of a significant drop $D_{i}$ is calculated by counting the number of drop points in a drop, i.e., $\lvert D_{i} \rvert$. Having defined the recovery rate of a significant drop, the recovery ability of the prediction framework is measured by the average of the recovery rate of all drops, given the performance result $\theta$. It is calculated using the average of the collected recovery rate. 
% According to the definition of the significant drop, the size of the drop is dependent on the moving average window size $M$. 
% The normalized recovery rate is designed to diminish the influence of moving average window size. 
The average recovery rate ($\mathbb{R}_{avg}$) of the performance result $\theta$ is defined as follows:

\[
\mathbb{R}_{avg} = \frac{\sum_{D_i \in \mathbb{D}(\theta)} \lvert D_i \rvert }{ |\mathbb{D}(\theta)|}
\]

% Normalized recovery rate ($\mathbb{R}_{norm}$):
% \[
% \mathbb{R}_{norm} = \frac{\mathbb{R}_{abs}}{M}
% \]
% If we would like to normalize, then 
% divide R_avg / length of theta

The predictive model has a better ability to recover from the drops with lower recovery rate values.

% \textcolor{red}{The absolute recovery rate of a performance drop $D = \langle p_u, \cdots, p_v\rangle$ is $v - u$.}

In essence, our proposed meta-measures enable decision-makers to assess the performance stability of predictive process models in event streams. Users can evaluate whether the predictive models exhibit a sufficiently stable performance for their specific business scenarios. 
Let us assume the example performance graphs by three different scenarios as in Fig.~\ref{fig:matrix}, where the average performances of the models are comparable. If the decision has to be made frequently or the process can be frequently and dynamically changed, the predictive model should be able to quickly adapt and recover the performance.  
% XL: I don't follow this:
In addition, the performance drops should preferably not deviate significantly from the moving average. This can be assessed through the recovery rate and the magnitude of performance drops, respectively; both should have a low value. 
For example, in scenarios 1 and 3, the recovery rate is relatively better (lower) than in Scenario 2, indicating a faster restoration of performance. Additionally, scenarios 1 and 3 exhibit smaller magnitudes of performance drops than scenario 2.
For business scenarios that require a high level of risk control, the predictive model should show less fluctuating performance to reduce the potential risk of incorrect predictions, i.e., having a low value on the frequency of performance drops and the volatility of performance. Scenarios 2 and 3 are better models for high-risk control than Scenario 1 due to the relatively smaller number of performance drops and a lower value on the volatility of the performance.

% \todo{Add some discussion wrt Fig. 3 here to close the link.}

\section{Empirical analysis and results}
\label{sec:empirical_analysis}
For the evaluation, we demonstrate the use of the proposed meta-measures for assessing and comparing the stability of different types of online frameworks. We implemented both the frameworks and the meta-measures. The code and data to reproduce the experiments presented in this paper are available at \url{https://github.com/ghksdl6025/online_ppm_stability}.
The objectives of the evaluation are threefold: (O1) demonstrate that the meta-measures \emph{assess} the stability of each individual prediction framework; (O2) use the meta-measures to \emph{compare} the stability of the frameworks; and (O3) \emph{select} a suitable framework for each of the three business scenarios. 

\paragraph{Datasets} For the evaluation, we consider four publicly accessible event logs. We reused two preprocessed and labeled event logs from the existing outcome prediction benchmark~\cite{teinemaa2019outcome}: the BPIC 2015\footnote{\scriptsize at: \url{https://data.4tu.nl/articles/dataset/BPI_Challenge_2015_Municipality_1/12709154/1}} and 
% is a log from a Dutch municipality of a process for granting building permissions. The outcome label in this log is 1 (true) when a trace contains the activity 'create procedure confirmation' and 0 otherwise. 
the BPIC 2017\footnote{\scriptsize at: \url{https://data.4tu.nl/articles/dataset/BPI_Challenge_2017_-_Offer_log/12705737}}. 
% event log refers to a personal loan request process at a Dutch financial institute. The outcome label evaluates to 1 (true) if a request is accepted, and 0 otherwise. 
The other two event logs are synthetic logs generated by processes in which concept drift has been applied. We consider the IRO5K and OIR5K\footnote{\scriptsize at: \url{https://data.4tu.nl/articles/dataset/Business_Process_Drift/12712436}} event log, synthetic logs regarding the assessment of loan applications~\cite{maaradji2016fast}. Both synthetic datasets, IRO5K and OIR5K, differ based on the type of concept drift in processes considered. The outcome label evaluates to 1 (true) if a request is accepted, and 0 otherwise. The two BPIC logs have been chosen because they are real-world event logs that have been used in the previous research on outcome predictive monitoring~\cite{teinemaa2019outcome}, and they differ greatly in terms of variability; Tab.~\ref{tab:event_log_statistics} lists some descriptive statistics of the four event logs. 
% XL: not sure needed 
%, the BPIC 2015 event log shows a higher number of activity labels and trace variants in respect of BPIC 2017. The IRO5K and OIR5K event logs have been chosen because they are characterized by process drift. The first 500 cases of both event logs follow the same process, whereas the subsequent cases are affected by two different types of process drifts that are generated by adding or removing events together with a loop condition. The process drift happens at the beginning and the end of the trace in OIR5K and IRO5K logs, respectively.

\begin{table}[]
\centering
\caption{Descriptive statistics of event logs used in the evaluation}
\label{tab:event_log_statistics}
\resizebox{\linewidth}{!}{%
\begin{tabular}{lrrrrrrrr}
\hline
 &
  \multicolumn{1}{c}{\textbf{\# cases}} &
  \multicolumn{1}{c}{\textbf{\# events}} &
  \multicolumn{1}{c}{\textbf{\# activity}} &
  \multicolumn{1}{c}{\textbf{\# variants}} &
  \multicolumn{1}{c}{\textbf{\begin{tabular}[c]{@{}c@{}}Avg \\ events/case \end{tabular}}} &
  \multicolumn{1}{c}{\textbf{\begin{tabular}[c]{@{}c@{}}Median \\ events /case\end{tabular}}} &
  \multicolumn{1}{c}{\textbf{\begin{tabular}[c]{@{}c@{}}\# true \\ labels\end{tabular}}} &
  \multicolumn{1}{c}{\textbf{\begin{tabular}[c]{@{}c@{}}\# false \\ labels\end{tabular}}} \\ \hline
BPIC 2015 & 1199 & 52217 & 289 & 1100 & 44 & 42 & 506 & 693  \\
BPIC 2017    & 1878 & 23941 & 22  & 376  & 13 & 15 & 576 & 1302 \\
IRO5K        & 1000 & 10756 & 20  & 111  & 11 & 12 & 237 & 763  \\
OIR5K        & 1000 & 11042 & 20  & 121  & 11 & 12 & 243 & 757  \\\hline
\end{tabular}%
}
\end{table}

\paragraph{Setup}
In this evaluation, we use three different types of online predictive models: (i) an incremental classifier, the Hoeffding adaptive tree classifier (HATC)~\cite{bifet2009adaptive,domingos2000mining}, (ii) a classical, light-weighted classifier, i.e., the eXtreme gradient boosting (XGB), and (iii) a deep learning model, i.e., Long Short-Term Memory (LSTM) neural network. We show that the proposed meta-measures can be used to assess and compare the stability of different types of predictive models.

% HATC, in particular, is a tree-based classifier, which incrementally constructs split points depending on the confidence for information gain. Any streaming classification framework that normally requires a grace period to initialize the model training regardless of the classifier~\cite{domingos2000mining}.
For the trace bucketing and encoding, we use the index-based encoding and consider a different maximum prefix length for each event log: 42 for BPIC 2015, 15 for BPIC 2017, and 12 for IRO5K and OIR5K as well, which corresponds to the median number of events per case in each log. The minimum prefix length is set to 2 for all event logs.

Regarding the training of the prediction frameworks, we use the first 200 labels received as a grace period in all experiments. Until the 200th label is observed, the events without labels in the stream are not processed, and the labels are used only to update the predictive models. 

After the grace period, each predictive model can be retrained and updated over time when the model training condition is triggered. Regarding the model updated policies, HATC is updated when a new label is observed, whereas XGB takes a sliding window~\cite{lee2022analysis}, which manages new labels, as a training set and has been trained from scratch. For LSTM, we train the model only once with the labels from the grace period without further updates. 

For the performance measure $\theta$, we used the accuracy, the recall, the precision, and the F1-score~\cite{scikit-learn}. For computing the moving average, we set $M$ to 30.

To implement the online predictive model, we used the Python package River~\cite{montiel2021river}, while the Scikit-learn~\cite{scikit-learn} and PyTorch~\cite{paszke2019pytorch} packages are used to implement the offline predictive models.

\paragraph{Results} 
In total, we run 3 [frameworks] * (41 + 14 + 11 + 11) [buckets] * 4 [performance measures, i.e., ACC, Recall, precision, f1] = 924 configurations where we calculated the four meta-measures.
We structure the experiment results with respect to the three evaluation objectives to ease the discussion.

\begin{table}[h]
\centering
\resizebox{\linewidth}{!}{%
\begin{tabular}{crrrrrr}
\hline
Prefix &
  \multicolumn{1}{c}{\begin{tabular}[c]{@{}c@{}}Average F1 score \\ from streaming\end{tabular}} &
  \multicolumn{1}{c}{Drops} &
  \multicolumn{1}{c}{\begin{tabular}[c]{@{}c@{}}Volatility of  \\ performance\end{tabular}} &
  \multicolumn{1}{c}{Max. Magnitude} &
  \multicolumn{1}{c}{Avg. Magnitude} &
  \multicolumn{1}{c}{Recovery rate} \\ \hline
2  & 0.536 & 54 & 0.074 & 0.418 & 0.115 & 8.556  \\
7  & 0.690 & 67 & 0.114 & 0.433 & 0.166 & 6.776  \\
14 & 0.860 & 31 & 0.037 & 0.396 & 0.068 & 13.194 \\ \hline
\end{tabular}%
}
\caption{F1-score and meta-measures of BPIC 2017 log with XGB classifier, for a short, medium, and long prefix length.}
\label{tab:meta_measures_prefix}
\end{table}

\begin{figure}[h]
    \centering
    \includegraphics[width=\linewidth]{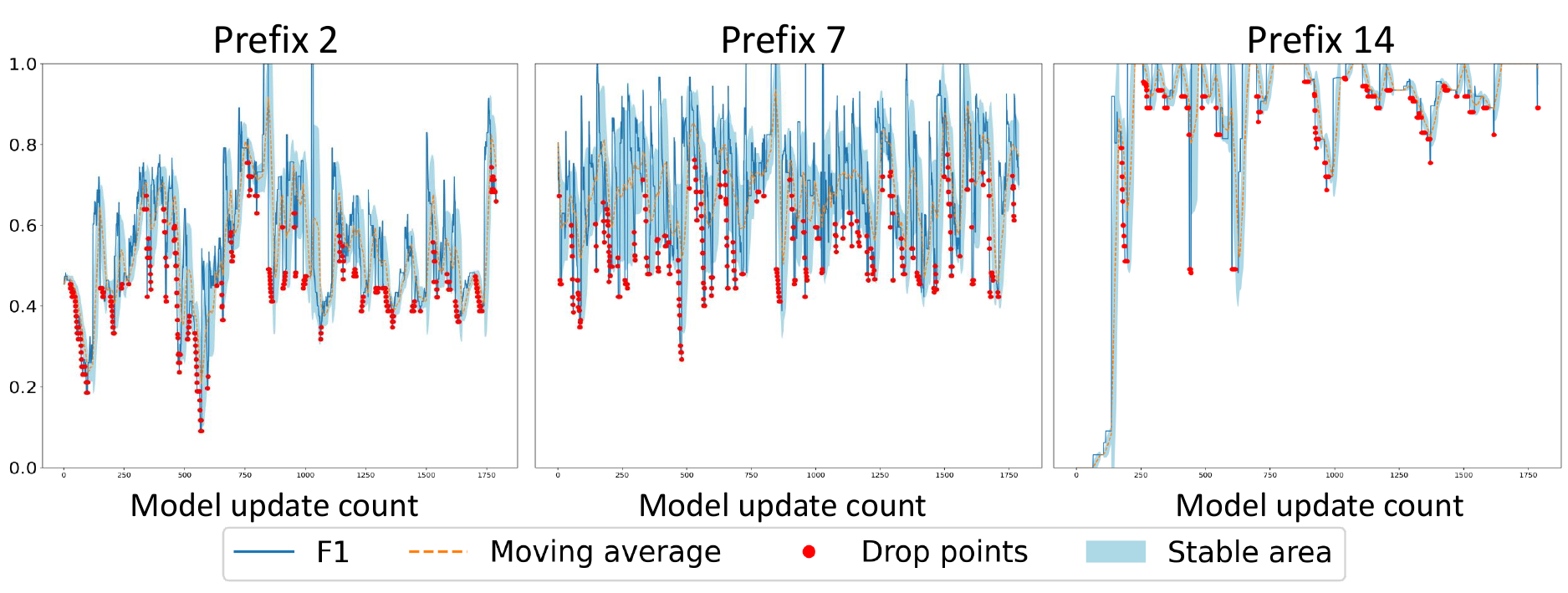}
    \caption{F1-score updates and performance drops for BPIC 2017 log with XGB classifier, for a short, medium, and long prefix length.}
    \label{fig:meta_measures_prefix}
\end{figure}

% The discussion of the experiment results is split into three parts. 
Firstly, we would like to show that (O1) the meta-measures can help to assess the stability of predictive models obtained with a given model at different prefix lengths, enriching the information given by the average F1-score over time. 

Table~\ref{tab:meta_measures_prefix} lists the average F1 scores and the values of the four meta-measures for the XGB model obtained on the BPIC17 log. We selected prefix lengths 2, 7, and 14, when using the BPIC17 log. Fig.~\ref{fig:meta_measures_prefix} shows the detailed moving average and drop points of the F1 score of the XGB model, for prefix lengths 2, 7, and 14, respectively.
% Fig.~\ref{fig:meta_measures_prefix} 
% shows for XGB, the average f1-score and the values of the four meta-measure obtained on the BPIC 2017 event log. 
As far as the average F1-score is concerned, the performance is higher at the longest prefix length, mainly due to the additional information available in the prefixes to train the predictive model. However, a longer prefix would not always guarantee an improvement in the prediction stability. For example, all the stability measures, except the recovery rate, increase, when the prefix length changes from 2 to 7, showing that the predictive model has become more unstable when the prefix length increases in this case. 

% Despite the improved average performance, the extra event information appears to increase the complexity of the training data, resulting in a more fluctuating performance. The prediction becomes more certain and stable with the longest prefix length. 
As indicated by the volatility of the performance, the XGB at prefix length 7 is much more volatile than the other two, with a value of 0.114, almost twice as much when compared to 0.074 at prefix length 2, and almost three times when compared to 0.037 at prefix length 14. The moving averages shown in Fig.~\ref{fig:meta_measures_prefix} confirm and demonstrate this high volatility visually. On the other hand, the recovery rates indicate that the XGB at prefix length 7 recovers much faster, while at prefix lengths 2 and 14, the XGB takes more time to recover to the moving average.

\begin{table}[t]

\begin{subtable}{1\linewidth}
\centering
    \resizebox{\linewidth}{!}{%
    \begin{tabular}{crrrrrr}
    \hline
    \multicolumn{1}{c}{\begin{tabular}[c]{@{}c@{}}Predictive \\ model\end{tabular}} &
      \multicolumn{1}{c}{\begin{tabular}[c]{@{}c@{}}Average F1 score \\ from streaming\end{tabular}} &
      \multicolumn{1}{c}{Drops} &
      \multicolumn{1}{c}{\begin{tabular}[c]{@{}c@{}}Volatility of  \\ performance\end{tabular}} &
      \multicolumn{1}{c}{Max. Magnitude} &
      \multicolumn{1}{c}{Avg. Magnitude} &
      \multicolumn{1}{c}{Recovery rate} \\ \hline
    HATC & 0.646 & 57 & \textbf{0.058} & \textbf{0.382} & \textbf{0.095} & \textbf{5.825} \\
    XGB  & \textbf{0.824} & 72 & 0.082 & 0.431 & 0.123 & 6.653 \\
    LSTM & 0.639 & \textbf{33} & 0.076 & 0.444 & 0.135 & 9.364 \\ \hline
    \end{tabular}%
    }
   \caption{Meta-measures: BPIC2017 log with prefix length 10}\label{tab:sub_first}
\end{subtable}

\bigskip
\begin{subtable}{1\linewidth}
\centering
    \resizebox{\linewidth}{!}{%
    \begin{tabular}{crrrrrr}
    \hline
    \multicolumn{1}{c}{\begin{tabular}[c]{@{}c@{}}Predictive \\ model\end{tabular}} &
      \multicolumn{1}{c}{\begin{tabular}[c]{@{}c@{}}Average F1 score \\ from streaming\end{tabular}} &
      \multicolumn{1}{c}{Drops} &
      \multicolumn{1}{c}{\begin{tabular}[c]{@{}c@{}}Volatility of  \\ performance\end{tabular}} &
      \multicolumn{1}{c}{Max. Magnitude} &
      \multicolumn{1}{c}{Avg. Magnitude} &
      \multicolumn{1}{c}{Recovery rate} \\ \hline
    HATC & 0.694 & 44 & \textbf{0.058} & \textbf{0.290} & \textbf{0.088} & 6.568 \\
    XGB  & \textbf{0.778} & 44 & 0.113 & 0.347 & 0.152 & \textbf{5.455} \\
    LSTM & 0.569 & \textbf{28} & 0.069 & 0.323 & 0.111 & 8.429 \\ \hline
    \end{tabular}%
    }
   \caption{Meta-measures: IRO5K log with prefix length 7}\label{tab:sub_second}
\end{subtable}
\caption{F1-score and meta-measures for the F1-score comparison by predictive models} \label{tab:meta_measures_model}
\end{table}

Second, (O2) we compare the stability of different predictive models for a given event log, to show that the proposed meta-measures can be used to compare the stability of different frameworks.

Table~\ref{tab:meta_measures_model} compares the performance and stability of the three considered models for the BPIC 2017 and IRO5K event logs. Among the three models, XGB outperforms the other two models in both event logs for the average F1-score from streaming. However, it is not the most stable predictive model according to the meta-measures. XGB has higher values on the meta-measures (more volatile) than the other two models. These findings from Tab.~\ref{tab:meta_measures_model} emphasize that if a stable predictive model is desired, which can recover quickly from performance drops albeit with slightly lower accuracy, HATC should be the preferred choice. As far as the performance drop is concerned, the deep learning model (LSTM) is also highly stable for the IROK5K log. However, LSTM appears not to have been trained well enough to achieve a high average performance (F1-score).

Finally, we show (O3) the selection of three configurations in Tab.~\ref{tab:configuration} that match the three critical business scenarios of Fig.~\ref{fig:matrix}. We used the results on the event logs with concept drift as an example. 

% \usepackage{graphicx}
\begin{table}[h]
\centering
\resizebox{\linewidth}{!}{%
\begin{tabular}{crrrrrr}
\hline
Configuration & \multicolumn{1}{c}{\begin{tabular}[c]{@{}c@{}}Average F1 score \\ from streaming\end{tabular}} & Drops & \multicolumn{1}{c}{\begin{tabular}[c]{@{}c@{}}Volatility of \\ performance\end{tabular}} & \multicolumn{1}{c}{Max. Magnitude} & \multicolumn{1}{c}{Avg. Magnitude} & \multicolumn{1}{c}{Recovery rate} \\ \hline
C1 & 0.69 & 44 & 0.058 & 0.290 & 0.088 & 6.568 \\
C2 & 0.95 & 26 & 0.018 & 0.096 & 0.032 & 7.692 \\
C3 & 0.94 & 24 & 0.024 & 0.251 & 0.041 & 11.042 \\ \hline
\end{tabular}
}
\caption{F1-score and meta-measures for the F1-score of three selected configurations for the demonstration}
\label{tab:configuration}
\end{table}

\begin{figure}[h]
    \centering
    \includegraphics[width=\linewidth]{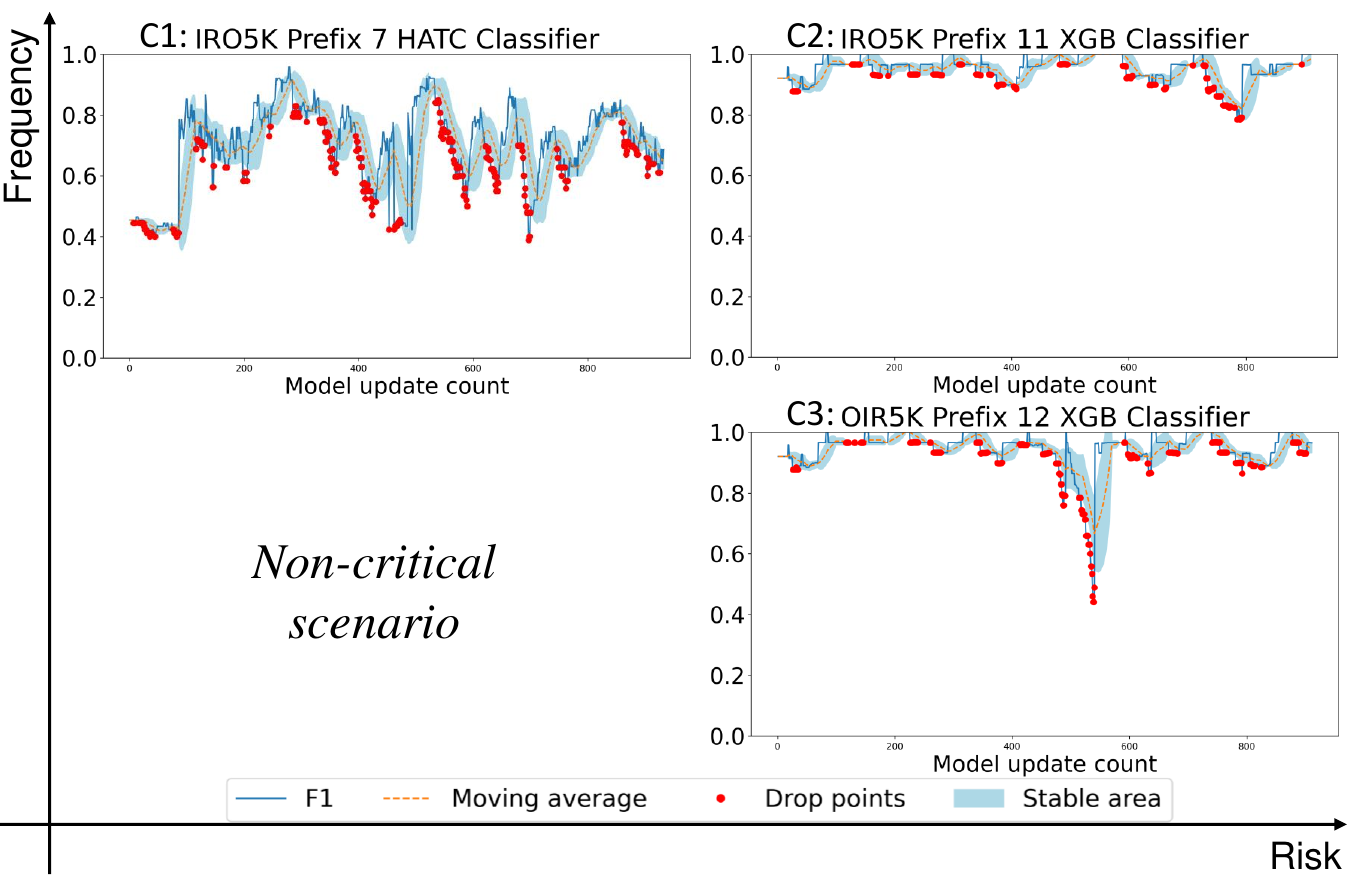}
    \caption{Three configurations selected based on the meta-measures, one for each critical scenario shown in Fig.~\ref{fig:matrix}}
    \label{fig:configuration}
\end{figure}

Figure~\ref{fig:configuration} illustrates three configurations, each tailored to one of the three critical business scenarios.  We use them to demonstrate how the proposed meta-measures can help to highlight the performance results that better fit each scenario. In terms of C1, the configuration exhibits higher volatility in performance, as indicated by the meta-measures, except for the recovery rate, which stands at 6.568. This high recovery rate makes C1 suitable for low-risk high-frequency scenarios. The next configuration, C2, has the lowest value of volatility of performance with a value of 0.018. Moreover, the other meta-measures of C2 are either the lowest or relatively lower compared to the other configurations, making it appropriate for high-risk high-frequency scenarios. For the high-risk low-frequency scenario, C3 can be considered. This configuration displays stable performance, characterized by the low value of performance drops and volatility. However, the recovery speed is comparatively slower than the other two configurations, measuring at 11.042, making it suitable for low-frequency scenarios. These findings illustrate how the proposed meta-measures aid in identifying performance results that align with specific business scenarios. 

\paragraph{Discussion and Limitations}
In the evaluation, we have shown that the meta-measures capture the stability of the performance in an intuitive way. For example, when the \emph{volatility of performance} is high, the performance is likely to be unstable. We have also shown that we can use the meta-measures to compare the stability of the performance.   

% In this paper, we provide some ``intuition'' behind the use of meta-measures for selecting a configuration for a scenario. We did not provide explicit methods. However, the ``intuition'' can be formalized and investigated systematically. For example, one may consider adding the weights to the meta-measures and building an automatic recommender. 

An important remark is that the meta-measures are defined using the moving average, which is computed based on the moving window size $M$. Thus, changing $M$ would have a large effect on the meta-measures. In this paper, we keep $M$ consistent across the configurations to be able to compare the meta-measures. In future work, we aim to investigate the change of $M$ and its influence on the meta-measures.

% Some other small issues. 
% - 

\section{Conclusions}
\label{sec:conclusion}

In this paper, we have proposed a performance measurement framework for outcome-oriented predictive process monitoring in online settings. The framework contains four performance meta-measures, which concern (1) the frequency of significant drops, (2) the volatility of performance, (3) the magnitude of performance drops, and (4) the recovery rate of the predictive model. Given a basic performance measure, i.e., any confusion matrix-derived performance measure for prediction, the proposed meta-measures can be used to derive additional performance measures. 

The application of the performance meta-measures allows us to characterize and evaluate the performance stability of an outcome-oriented predictive process monitoring model in online settings. These meta-measures serve as valuable tools to assess the appropriateness of a predictive model in various business scenarios that differ in terms of decision-making frequency and risk. For instance, a business scenario characterized by a high frequency with low-risk decision-making will require a predictive process outcome model with a minimum magnitude of significant performance drops and, most importantly, quick recovery to an acceptable performance level after such drops. By employing performance meta-measures, we can determine the suitability of predictive models for different business contexts and make informed decisions accordingly.

For future work, the proposed framework can be extended with various performance evaluation methods. For example, instead of the prefix-length perspective, we may also complement the framework with a method that evaluates the accuracy from the last-state perspective. In addition to our in-depth exploration of the framework, we also intend to conduct an analysis aimed at uncovering the underlying causes behind instances of performance drops. Providing performance evaluation from multiple perspectives may provide comprehensive knowledge and explainability of online predictive monitoring models, which also merits further investigation.

\bibliographystyle{IEEEtran}
\bibliography{mybiblio}

\end{document}